\documentclass[10pt,twocolumn,letterpaper]{article}

\usepackage[datasets]{wacv}             

\usepackage{graphicx}
\usepackage{amsmath}
\usepackage{amssymb}
\usepackage{booktabs}
\usepackage{array}
\usepackage{multirow}
\setcitestyle{numbers,sort}

\definecolor{wacvblue}{rgb}{0.21,0.49,0.74}
\usepackage[pagebackref,breaklinks,colorlinks,allcolors=wacvblue]{hyperref}

\def\wacvPaperID{3762} 
\def\confName{WACV}
\def\confYear{2027}

\title{Auditing Metrics for Synthetic Data Utility: A Cross-Domain Study of Appearance and Geometry}

\author{
Patryk Bartkowiak$^{1,4}$ \quad
Dominik L. Michels$^{2,4}$ \quad
S\"oren Pirk$^{3,4}$ \quad
Wojciech Pa{\l}ubicki$^{1,4}$\\
$^{1}$Adam Mickiewicz University \quad
$^{2}$KAUST \quad
$^{3}$Kiel University\\
$^{4}$GreenMatterAI
}

\newcommand{\bodyfigw}{1.0}
\newcommand{\teaserpanel}{0.32}

\begin{document}
\maketitle


\begin{abstract}
Image-similarity metrics are widely used to assess synthetic data, yet it remains unclear whether they predict how effectively synthetic training data transfer to real images.
We present a common evaluation protocol for appearance- and structure-oriented proxies and apply it to seven benchmark families, 21 dataset variants, and 87{,}880 image-pair comparisons spanning object detection, semantic segmentation, and 6D pose estimation.
No standalone metric is consistently strongest: appearance metrics are more informative in settings dominated by rendering, texture, or illumination differences, whereas correspondence-based metrics provide stronger evidence when pose, viewpoint, or spatial structure varies.
These results show that the two signal families capture distinct aspects of synthetic-to-real compatibility. As a secondary analysis, we introduce SADGE, a transparent three-parameter interaction between DINOv3 appearance similarity and MASt3R correspondence.
When fitted and evaluated on the complete benchmark, SADGE reaches Pearson \(r=0.834\) and Spearman \(\rho=0.853\), compared with \(r=0.585\) and \(0.545\) for the strongest appearance- and structure-oriented standalone metrics.
Keeping DINOv3 and MASt3R fixed, we repeatedly fit SADGE on subsets of variants and evaluate it on the remaining variants.
Median held-out Pearson correlation increases from \(0.611\) with three calibration variants to \(0.755\) with ten, exceeding the strongest unfitted metric from five calibration variants onward.
Under exhaustive family-disjoint calibration, median held-out correlation rises from \(0.529\) with two calibration families to \(0.841\) with five, although the wider dispersion and negative tail reveal substantial sensitivity to calibration composition.
Together, these analyses support evaluating appearance and structural compatibility jointly.
SADGE requires calibration on known downstream results; scoring additional datasets then requires no downstream training.

\end{abstract}

\providecommand{\bodyfigw}{1.0}
\begin{figure*}[t]
  \centering
  \includegraphics[width=\bodyfigw\linewidth]{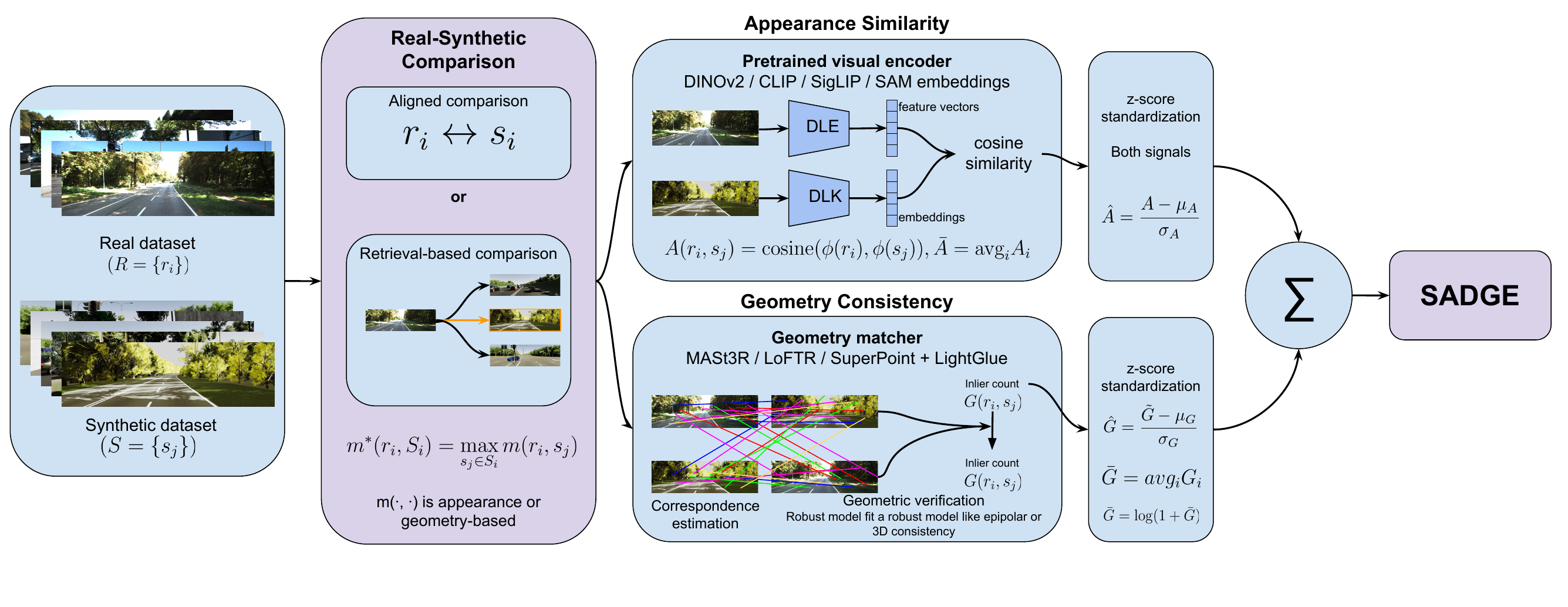}
  \vspace{-10mm}
  \caption{
    Our evaluation measures the association between synthetic-data utility and two complementary signal families: \emph{appearance similarity} and \emph{geometry consistency} between real and synthetic images.
    For each real image, comparison is performed either using an aligned real--synthetic pair or by retrieving the best synthetic match from a candidate subset.
    We benchmark each signal independently; as a secondary analysis, a constrained bilinear interaction called SADGE probes whether the two families contain complementary information on the assembled benchmark.
    }
    \vspace{-4mm}
  \label{fig:overview}
\end{figure*}

\section{Introduction}
\label{sec:intro}

Synthetic data is widely used to scale vision systems when real annotations are expensive or difficult to obtain, particularly for edge cases~\cite{schieber2024indoor-synthetic-review,lu2023ml-synthetic-data-review}.
Yet a basic measurement question remains unresolved: \emph{which properties of a synthetic dataset indicate that it will be useful for learning on real images?}

Evaluating synthetic data is still largely a trial-and-error process.
Practitioners generate candidate datasets, train models, and measure performance on held-out real data to determine whether a rendering pipeline, domain-randomization strategy, or generative process is effective.
In industrial and safety-critical settings, this loop is particularly costly: seemingly minor changes in illumination, background, material appearance, or object placement can substantially affect transfer performance~\cite{eversberg2021pbr-realism-vs-domain-rand,zhu2023towards-sim2real-industrial-parts,zhu2024automated-assembly-quality-inspection,horvath2022sim2real-domain-randomization-robotics}.
This motivates inexpensive proxy metrics, but a proxy is useful only if its relationship to downstream transfer has been established.

Recently, the need for such an estimate has only grown as synthetic data generation has diversified. Classical graphics-based pipelines render large datasets from simulators and CAD assets~\cite{greff2022kubric,martinezgonzalez2021unrealroxplus,kar2019metasim,raistrick2023infinigen}, offering explicit control over geometry, camera pose, lighting, materials, and sensor effects. In parallel, image-generative models synthesize training data by sampling from learned image distributions~\cite{zhang2021datasetgan,rombach2022ldm}, often increasing semantic and stylistic diversity but without explicit guarantees of geometric consistency. 
%
Common proxies include CLIP similarity, FID, LPIPS, PSNR, and DINO embeddings~\cite{li2025benchmarking,zenith2025sdqm,ko2022synbench}.
Specialized training-free metrics such as CLER~\cite{li2025benchmarking} use CLIP-derived, class-centered representations for classification-oriented data relevance.
These choices predominantly describe appearance, and they are often reported with an implicit assumption that better visual similarity implies better transfer.
What is missing is a common evaluation that asks how these existing metrics relate to downstream utility across different synthetic sources, target domains, and vision tasks, and whether structure-oriented correspondence provides a distinct signal.

This paper addresses that evaluation gap.
We assemble a diverse benchmark of published transfer results, including six real-to-synthetic VKITTI2 variants, and evaluate twelve commonly used appearance- and structure-oriented metrics under one aligned/retrieval-based comparison protocol.
The purpose is not to establish a universally optimal predictor, but to identify which measurable signals are informative, where they fail, and what their comparison reveals about synthetic-data utility.
Our central empirical finding is that the answer is not appearance \emph{or} geometry: the strongest standalone metric changes across domain-gap mechanisms, while dense geometric correspondence contributes information that appearance embeddings do not capture.

As a secondary exploratory analysis, we introduce SADGE (\textit{Structural and Appearance Domain Gap Estimator}), a low-capacity fusion of dataset-level appearance and geometric similarity.
SADGE is a three-parameter diagnostic probe. Its role is to test whether the two signal families contain complementary information on the assembled benchmark, not to establish that a fixed calibration predicts utility in an unseen domain.

\providecommand{\teaserpanel}{0.49}
\begin{figure*}[tp]
  \centering
  
  \includegraphics[width=\teaserpanel\linewidth]{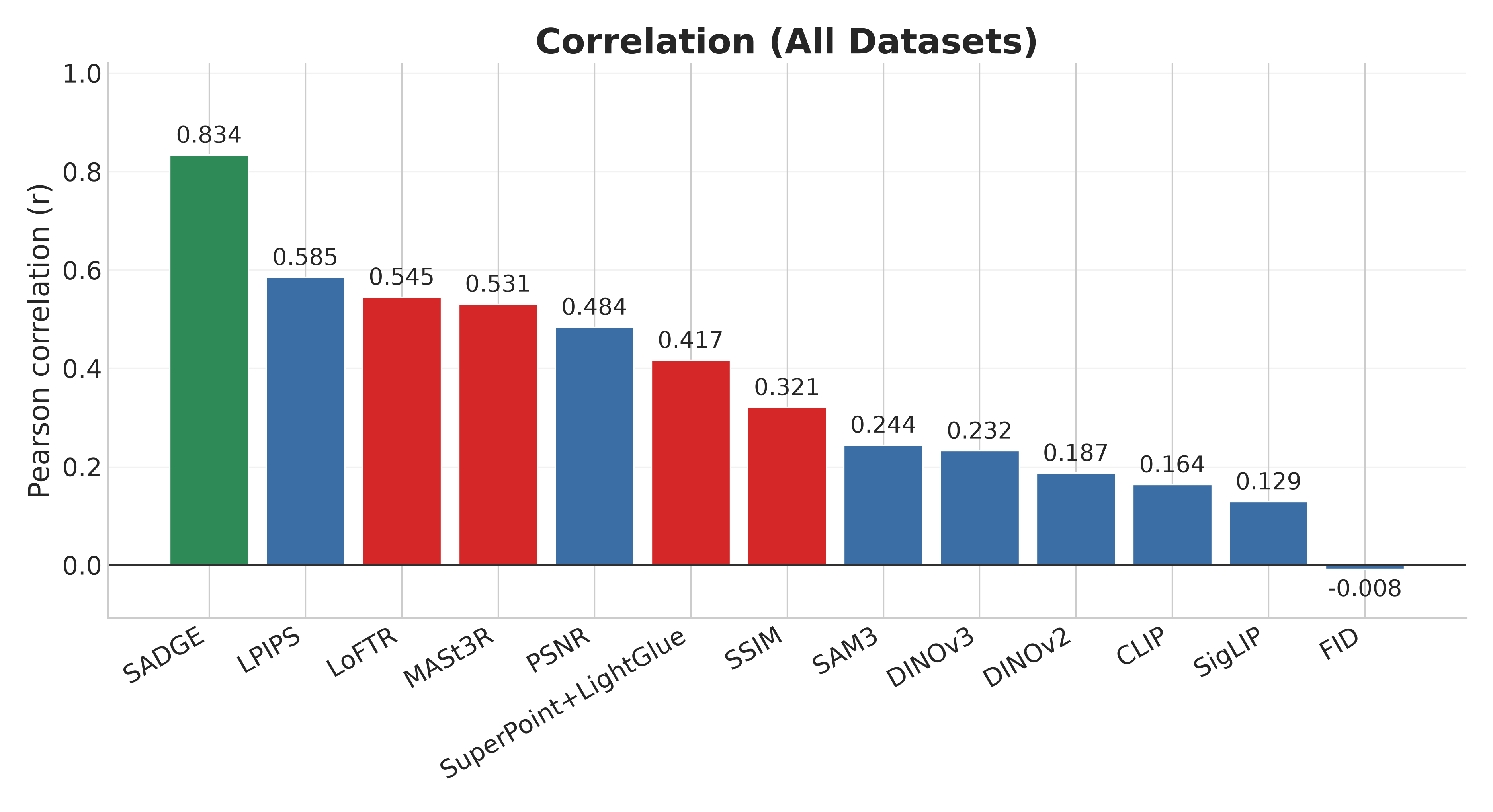}
  \includegraphics[width=\teaserpanel\linewidth]{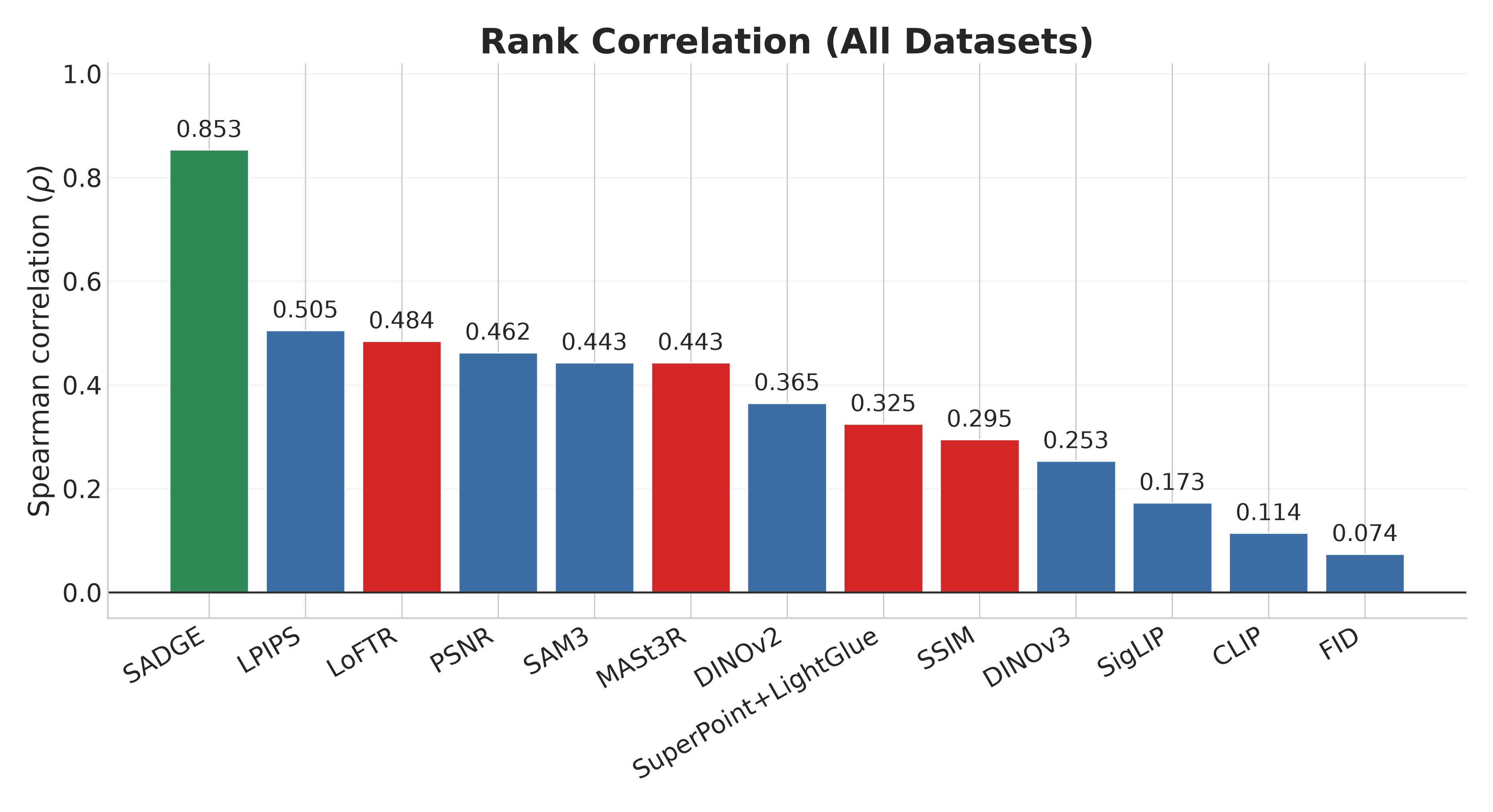}
  \includegraphics[width=\teaserpanel\linewidth]{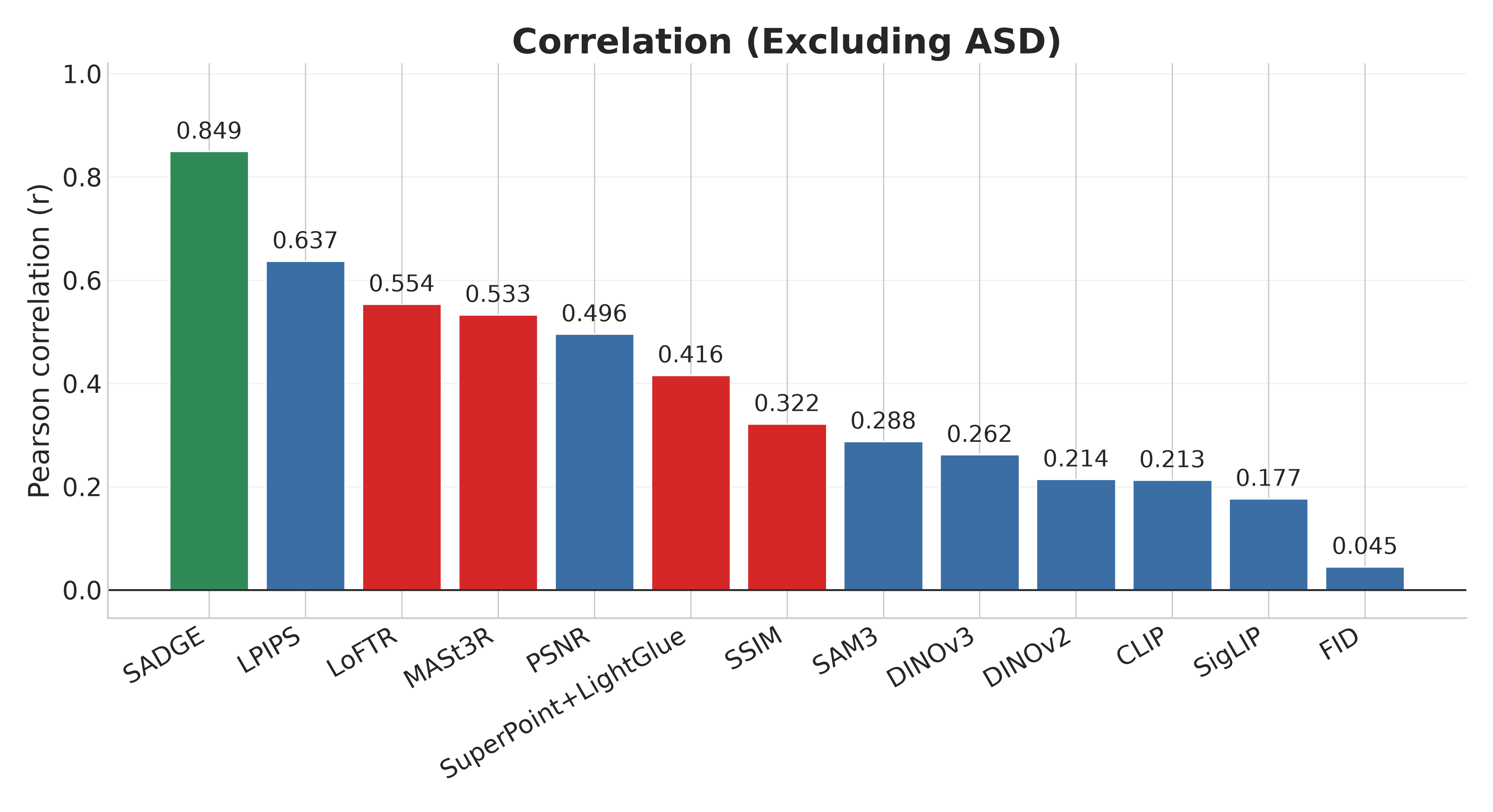}
  \includegraphics[width=\teaserpanel\linewidth]{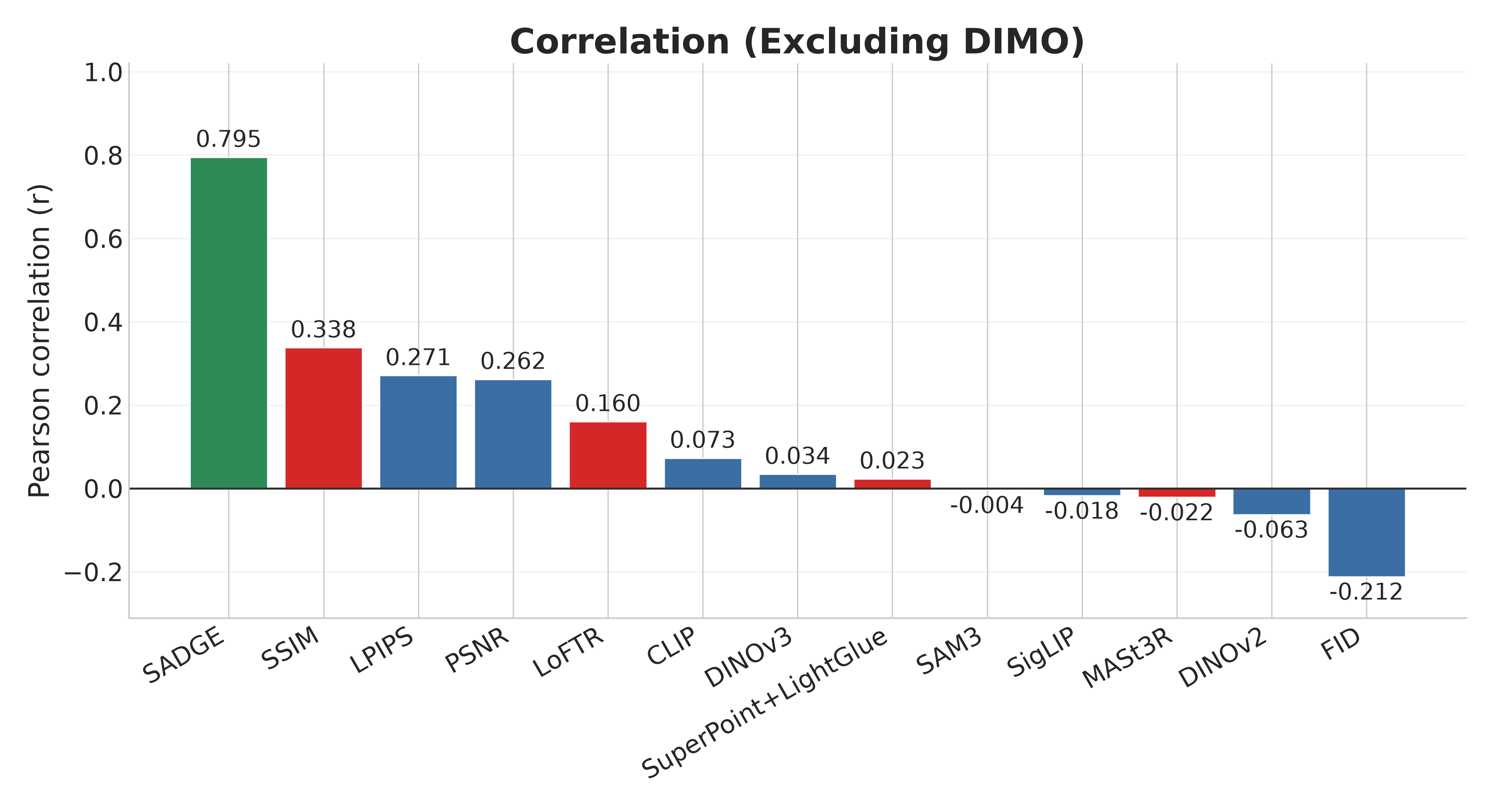}
  \includegraphics[width=\teaserpanel\linewidth]{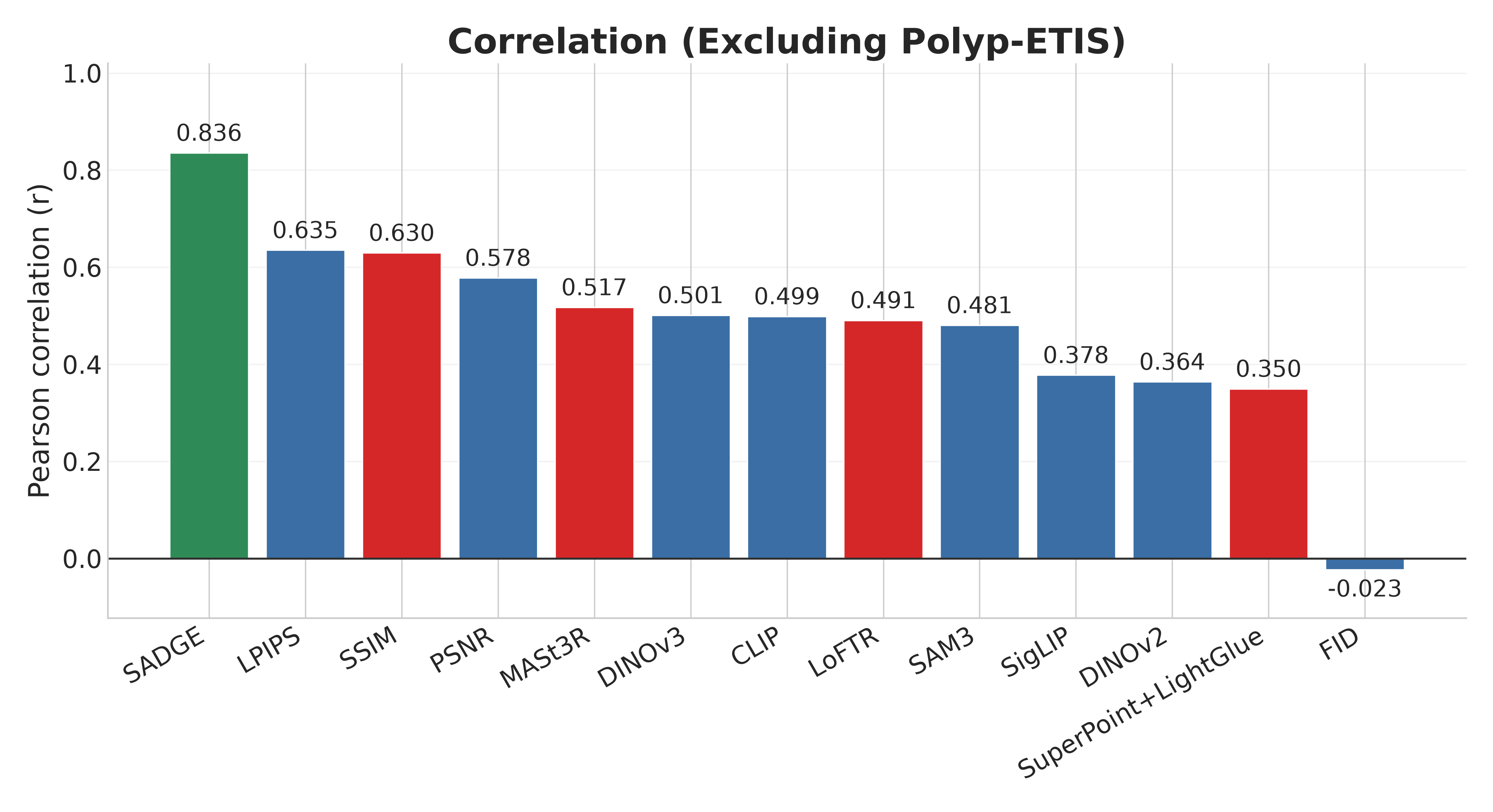}
  \includegraphics[width=\teaserpanel\linewidth]{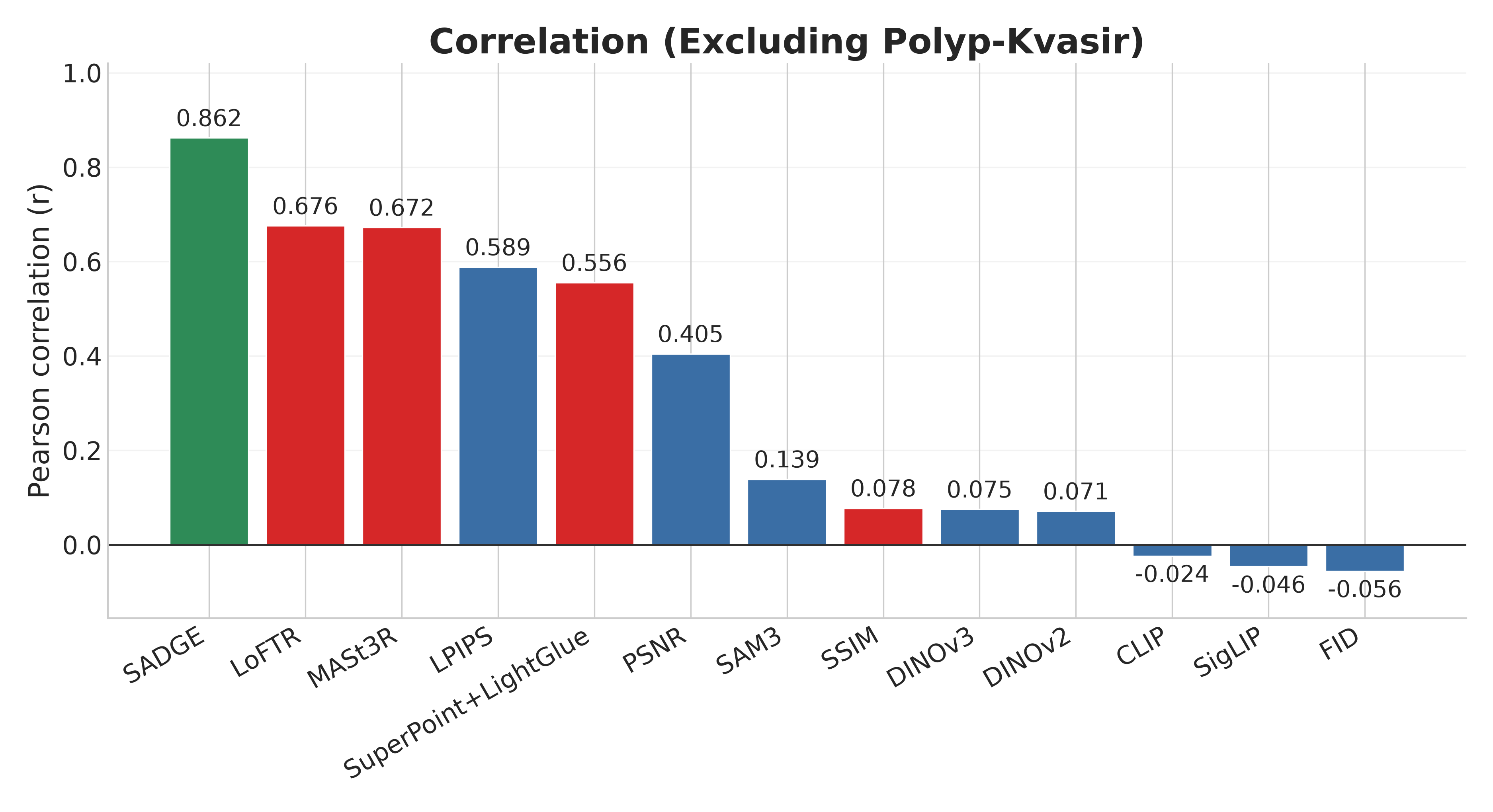}
  \includegraphics[width=\teaserpanel\linewidth]{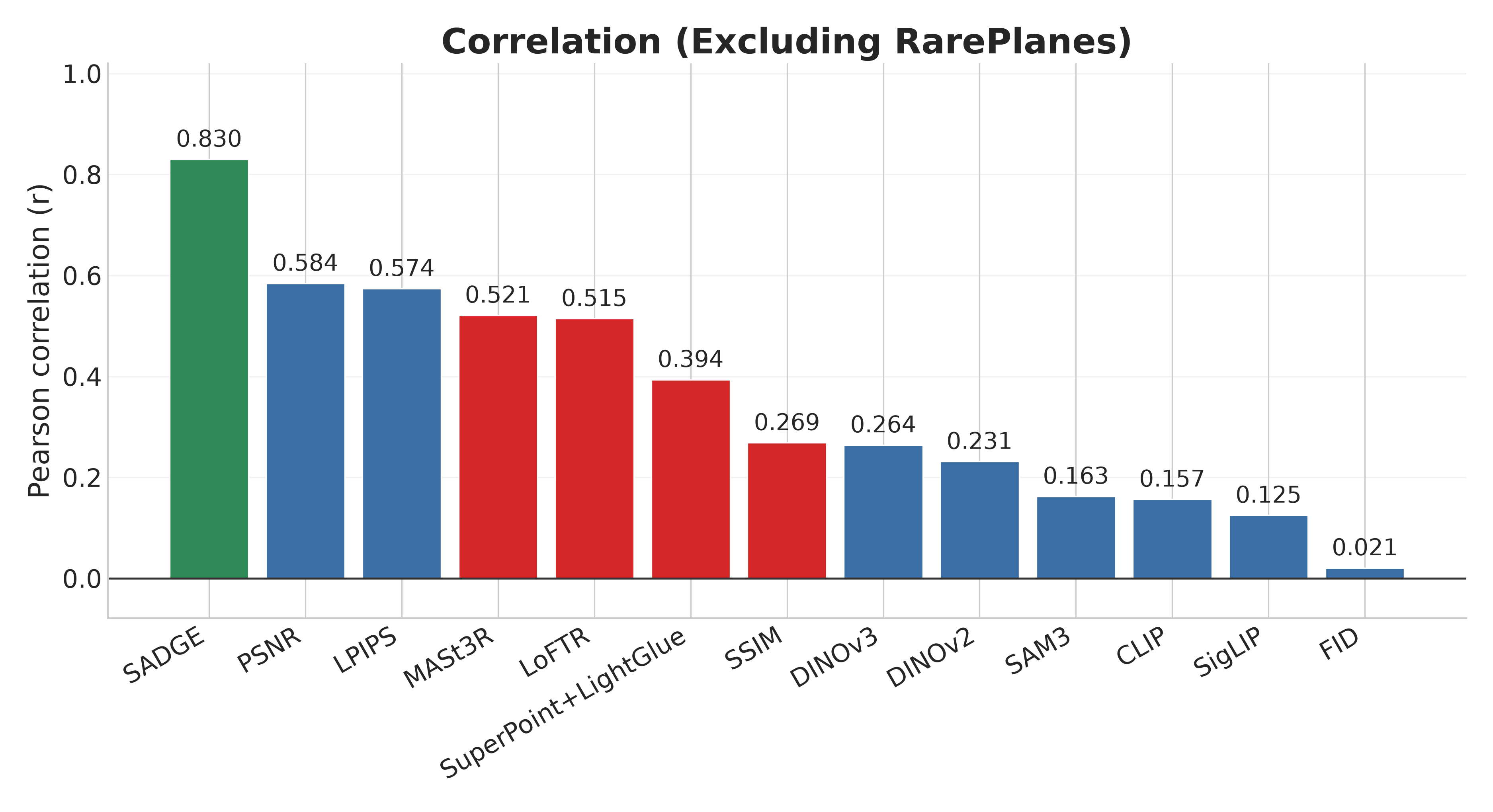}
  \includegraphics[width=\teaserpanel\linewidth]{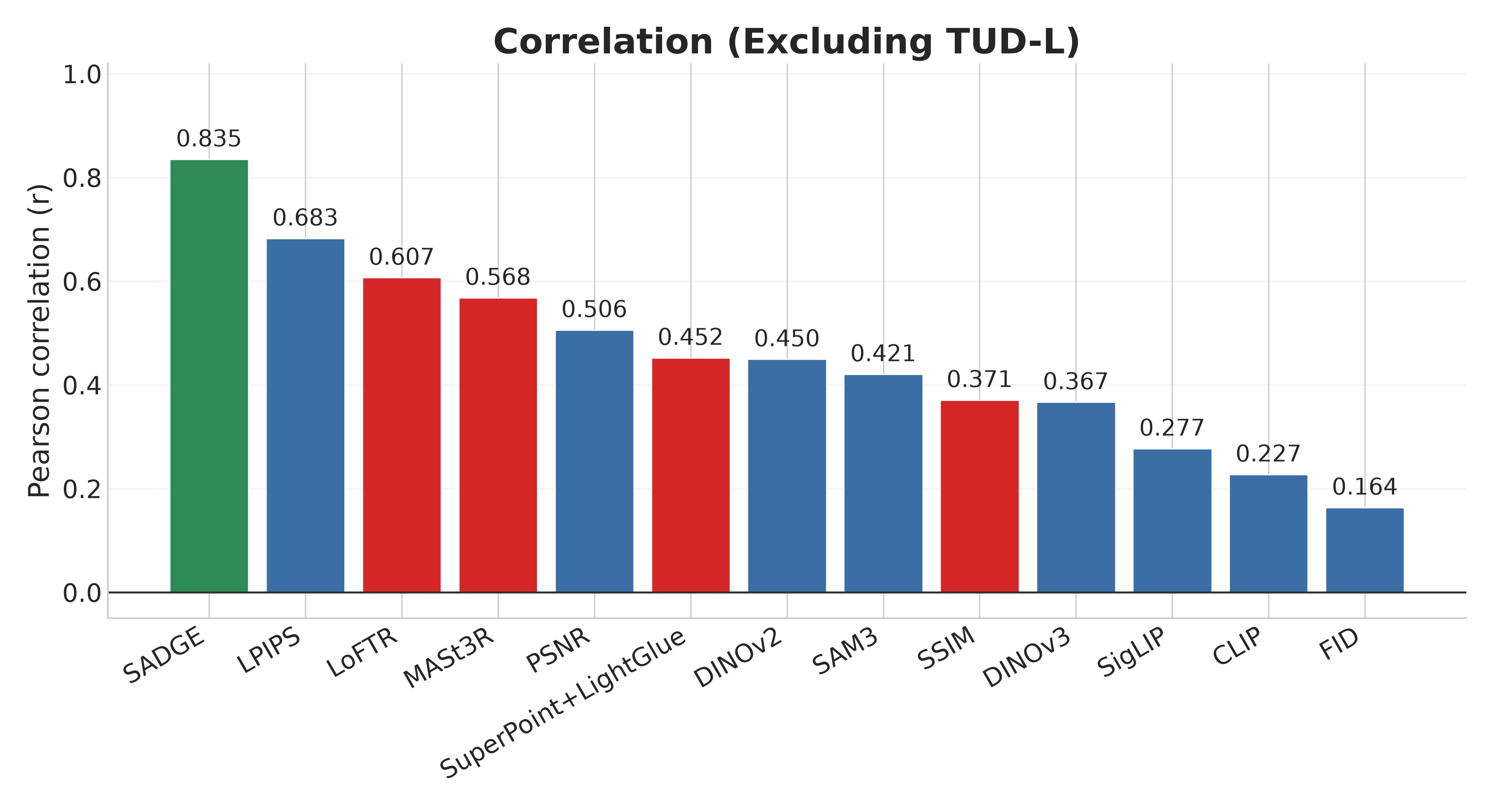}
  \includegraphics[width=\teaserpanel\linewidth]{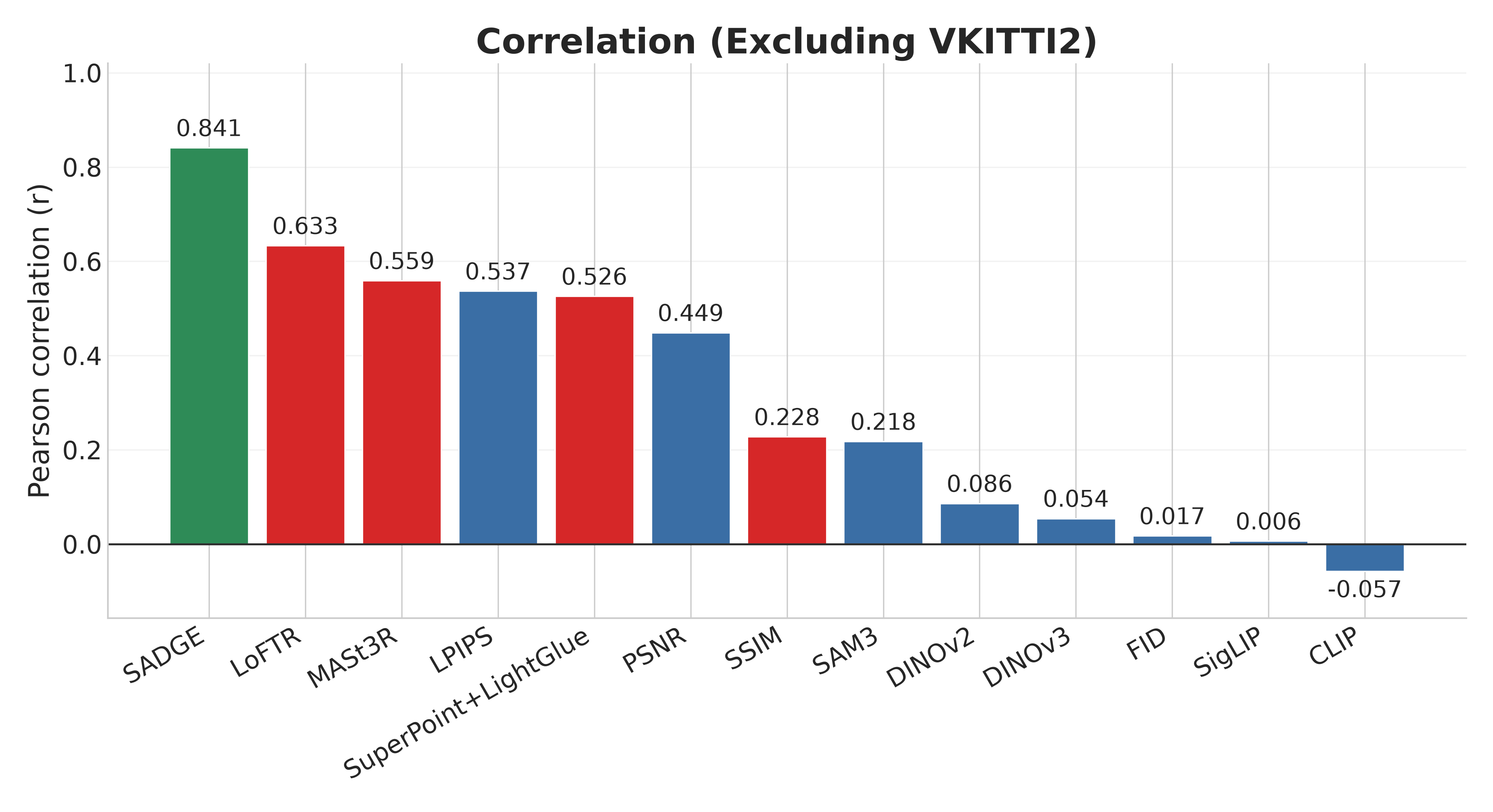}
  
  \vspace{-4mm}
  \caption{
    Correlation with downstream performance over all 21 dataset variants (first panel, Pearson \(r\); second panel, Spearman \(\rho\)) and Pearson correlation under family-exclusion sensitivity analysis (remaining panels: excluding ASD/agricultural, DIMO, Polyp-ETIS, Polyp-Kvasir, RarePlanes, TUD-L, and VKITTI2).
    Bar colors denote metric families: green is SADGE (ours), red is structure-oriented metrics (MASt3R, LoFTR, SuperPoint+LightGlue, and SSIM), and blue is appearance-oriented metrics (LPIPS, PSNR, CLIP, SigLIP, DINOv2/DINOv3, SAM3, and FID). Correlations are plotted signed, so a metric that is anti-correlated with downstream performance on a split appears below the zero line.
    The best standalone proxy changes with benchmark composition, illustrating that appearance and structure respond to different domain-gap mechanisms. The exploratory appearance--geometry summary ranks first in every panel (overall \(r=0.834\) and \(\rho=0.853\); family-exclusion \(r \in [0.795, 0.862]\)).
    }
    \vspace{-4mm}
  \label{fig:teaser}
\end{figure*}

We evaluate seven benchmark families that expose different synthetic-data failure modes: illumination shifts in DIMO~\cite{deroovere2022dimo}, weather and appearance variation in Virtual KITTI2~\cite{cabon2020vkitti2}, industrial object rendering for 6D pose estimation in TUD-L~\cite{hodan2017tless,hodan2018bop}, aerial synthetic-to-real transfer in RarePlanes~\cite{shermeyer2021rareplanes}, agricultural transfer in ASD~\cite{cieslak2024generating}, and generative polyp segmentation against Kvasir-SEG and ETIS-LaribPolypDB~\cite{dorjsembe2024polypddpm,jha2020kvasirseg,silva2014etis}.
The collection contains 21 dataset-level variants and 87{,}880 real--synthetic pair comparisons across object detection, semantic segmentation, and 6D pose estimation.

Our contributions are, in order:
\begin{enumerate}
    \item We provide a systematic cross-domain evaluation of twelve existing appearance- and structure-oriented proxies for synthetic-data utility, together with a unified protocol for aligned and unpaired real--synthetic data.
    \item We identify a consistent qualitative pattern across the study: no single proxy family dominates every setting, the useful signal changes with the domain-gap mechanism, and dense geometry matchers add information beyond appearance similarity.
    \item We document the benchmark construction, pairing protocol, metric families, runtime, and asset provenance to provide an auditable basis for continued study of synthetic-data utility.
    \item As a secondary analysis, we use SADGE, a transparent low-capacity joint score, to probe complementarity between appearance and geometry on the assembled benchmark.
\end{enumerate}

The metric evaluation and cross-domain analysis are the core contributions.
The SADGE experiment makes the observed structure--appearance relationship concrete, but acceptance of the study does not depend on treating its fitted coefficients as a general-purpose predictor.

\section{Related Work}
\label{sec:related}

We review prior work on synthetic-data quality metrics, representation-based proxies, geometric correspondence, and dataset interpretability.
Our focus is not only a new score, but the missing comparative question: how strongly do these existing signal families track synthetic-to-real utility when evaluated under a common protocol?
%
Common image similarity measures such as PSNR and SSIM quantify pixel-level fidelity or structural similarity~\cite{turaga2004psnr,wang2004ssim}. Learned perceptual metrics like LPIPS improve perceptual alignment~\cite{zhang2018lpips}. In generative modeling, the Inception Score (IS) and Fr\'echet Inception Distance (FID) are widely used to assess realism and diversity~\cite{salimans2016is,heusel2017fid}, yet they are known to have limitations and sensitivity to evaluation setup~\cite{borji2022gan-eval}. More recent distributional metrics include MAUVE~\cite{pillutla2023mauve} and precision--recall style scores for generative models~\cite{kynkaanniemi2019precision-recall}, and Authenticity~\cite{alaa2022faithful}. While these metrics are valuable, they do not directly indicate whether a synthetic dataset will improve a downstream model.
%
Closest to our evaluation objective is CLER~\cite{li2025benchmarking}, a training-free metric for classification-oriented generated or real data using CLIP-based, class-centered representations.
A second relevant line is SDQM~\cite{zenith2025sdqm}, which targets synthetic-data quality but relies on downstream training signals and remains largely appearance-centric in its formulation.
CLER demonstrates that generic appearance metrics can correlate weakly with downstream classification accuracy.
Our study broadens the empirical question to synthetic-to-real object detection, segmentation, and pose estimation and, crucially, evaluates geometric correspondence alongside appearance similarity.
Recent work has adopted pretrained representations as proxies for data usefulness.
Common embeddings include CLIP~\cite{radford2021clip}, SigLIP~\cite{zhai2023siglip}, DINOv2~\cite{oquab2023dinov2}, and newer foundation models such as DINOv3~\cite{simeoni2025dinov3}, as well as segmentation-centric representations like SAM~\cite{kirillov2023sam}.
For geometry-aware comparison, local and dense matchers such as SuperPoint~\cite{detone2018superpoint} with LightGlue~\cite{lindenberger2023lightglue}, LoFTR~\cite{sun2021loftr}, and MASt3R~\cite{leroy2024mast3r} have demonstrated strong correspondence quality.
Although these methods were not designed as synthetic-data utility metrics, they let us test whether structural agreement explains transfer behavior that appearance proxies miss.
Beyond distributional scores, dataset interpretability methods aim to quantify the value of individual examples or subsets. $\mathcal{V}$-usable information characterizes dataset difficulty with respect to a model family~\cite{ethayarajh2022vusable}, while Data Maps analyze training dynamics to identify easy, ambiguous, and hard examples~\cite{swayamdipta2020datamaps}. These perspectives motivate metrics that connect data characteristics to task performance.
%
Synthetic datasets have been used to study sim-to-real transfer in specialized domains, including aerial imagery and object detection~\cite{shermeyer2021rareplanes,deroovere2022dimo}. Synthetic datasets also provide high-quality labels for many vision tasks, including semantic and instance segmentation, text localization, object detection, and classification. Large-scale synthetic corpora such as CLEVR~\cite{johnson2017clevr}, ScanNet~\cite{dai2017scannet}, SceneNet RGB-D~\cite{mccormac2017scenenet}, NYU Depth v2~\cite{silberman2012nyu}, SYNTHIA~\cite{ros2016synthia}, KITTI~\cite{10.1177/0278364913491297}, Virtual KITTI2~\cite{cabon2020vkitti2}, and FlyingThings3D~\cite{mayer2016flyingthings} are widely used for task-specific benchmarking. Yet fixed datasets often lack all annotation types (e.g., camera pose, flow, or dense masks) and can introduce dataset biases~\cite{torralba2011bias,azulay2019transform}. In contrast, recent generators like Kubric can synthesize multiple cues per scene under diverse viewpoints and lighting~\cite{greff2022kubric}. These limitations motivate a comparative evaluation across synthetic sources rather than conclusions drawn from one metric or one dataset.

\providecommand{\bodyfigw}{1.0}
\begin{figure*}[t]
  \centering
  \includegraphics[width=\bodyfigw\linewidth]{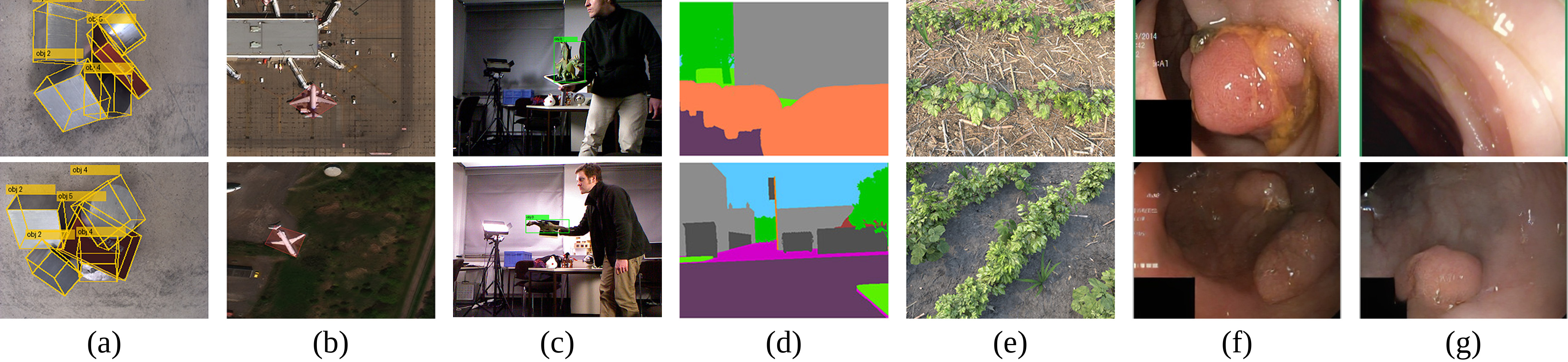}
  \vspace{-6mm}
  \caption{The seven benchmark families, with two examples each: DIMO (a), RarePlanes (b), TUD-L (c), VKITTI2 (d), ASD (e), Polyp-Kvasir (f), and Polyp-ETIS (g). The examples include image-only views, annotation overlays, and annotation-only visualizations, and span industrial, aerial, and medical domains, synthetic and real imagery, and diverse factors such as lighting, weather, background clutter, and viewpoint changes. The supplementary material shows the remaining synthetic sources of the two polyp families.}
  \vspace{-4mm}
  \label{fig:datasets}
\end{figure*}
\section{Evaluation Protocol}
\label{sec:method}


Our primary objective is to compare existing proxies for synthetic-data utility under a common protocol.
We organize them into appearance- and structure-oriented families, apply the same aligned or retrieval-based real--synthetic comparison, and aggregate each metric to one value per dataset variant.
This provides the basis for asking which standalone signals track downstream transfer and whether their behavior changes with the source of the domain gap.

Let $\mathcal{R}=\{r_i\}_{i=1}^{N}$ and $\mathcal{S}=\{s_j\}_{j=1}^{M}$ denote the real and synthetic image sets.

\subsection{Real--Synthetic Image Comparison}
\label{sec:image_comparison}

We compare each real image $r_i \in \mathcal{R}$ against one or more synthetic candidates from $\mathcal{S}$. Depending on the dataset we either use an aligned or retrieval-based comparison to identify image pairs.

\textbf{Aligned comparison.}
When paired real--synthetic samples are available, each real image $r_i$ is matched to a predefined synthetic counterpart $s_i$. This is the case for synthetic datasets which have been generated to closely reproduce the viewpoint, scene content, or layout of real data (e.g., such as in VKITTI2).

\textbf{Retrieval-based comparison.}
When exact pairs are not available, we search over a candidate synthetic subset and retain the best match for each real image according to the metric under consideration.
For computational efficiency, we do not compare against all $M$ synthetic images.
Instead, for each real image $r_i$ we uniformly sample a subset $\mathcal{S}_i \subset \mathcal{S}$ with $|\mathcal{S}_i| = k$.
The sampled candidate indices are cached and shared by all pairwise metrics, so every metric sees the same candidates even though it may select a different best match.
We orient every pairwise score so that larger values mean greater similarity: cosine similarity, PSNR, SSIM, and verified inlier counts are used directly, while LPIPS is negated.
Let $m(\cdot,\cdot)$ denote one such higher-is-better pairwise score.
The retrieval-based score for $r_i$ is
\begin{equation}
m^*(r_i,\mathcal{S}_i) = \max_{s_j \in \mathcal{S}_i} m(r_i,s_j).
\end{equation}
Because FID is a distribution-level distance, it is the exception to Eq.~(1): FID is computed once between the evaluated real and synthetic image sets for a dataset variant and is negated when correlations are reported.
The exploratory joint score combines independently aggregated appearance and geometry measurements; it does not require the two metrics to select the same synthetic image.

\subsection{Appearance- and Structure-Oriented Metrics}
\label{sec:appearance_similarity}

To quantify visual similarity, we compute an appearance similarity score in a learned feature space. Let $\phi(\cdot)$ denote a visual encoder that maps an image to a semantic feature representation. The appearance similarity between a real image $r_i$ and a synthetic image $s_j$ is defined as
\begin{equation}
A(r_i,s_j) = \frac{\phi(r_i)^\top \phi(s_j)}
{\|\phi(r_i)\|_2 \, \|\phi(s_j)\|_2},
\end{equation}
i.e., the cosine similarity between normalized image embeddings.

For the joint SADGE instantiation we use DINOv3, while the evaluation also includes CLIP, SigLIP, DINOv2, SAM3, LPIPS, PSNR, and FID.
The dataset-level appearance score is the mean over the selected pairs, $\bar{A} = \tfrac{1}{N} \sum_{i=1}^{N} A_i$, where $A_i$ denotes either the aligned or retrieval-based appearance similarity score for image~$r_i$.

\begin{table*}[t]
\centering
\small
\caption{Metrics evaluated under the common protocol. All reported scores are oriented so that larger values indicate greater real--synthetic similarity. FID is evaluated at dataset level and therefore does not participate in per-image retrieval.}
\label{tab:metric_protocol}
\setlength{\tabcolsep}{4pt}
\begin{tabular}{p{2.5cm}p{3.6cm}p{2.1cm}p{8.0cm}}
\toprule
Signal family & Metrics & Evaluation level & Orientation and aggregation \\
\midrule
Appearance embeddings & CLIP, SigLIP, DINOv2, DINOv3, SAM3 & Image pair & Cosine similarity; maximize over candidates, then mean over real images \\
Perceptual / pixel & LPIPS & Image pair & Negative distance; maximize (equivalently minimize LPIPS), then mean \\
Perceptual / pixel & PSNR & Image pair & Maximize over candidates, then mean \\
Distributional & FID & Dataset variant & Negative FID between real and synthetic evaluated sets \\
Local structure & SSIM & Image pair & Maximize over candidates, then mean \\
Geometric correspondence & SuperPoint+LightGlue, LoFTR, MASt3R & Image pair & Verified inlier count; maximize over candidates, then mean \\
\bottomrule
\end{tabular}
\end{table*}


Appearance similarity captures semantic and textural proximity, but not necessarily structural agreement.
We therefore evaluate structure-oriented metrics as a separate family: SSIM and the correspondence matchers MASt3R, LoFTR, and SuperPoint+LightGlue.
Given a real--synthetic pair $(r_i,s_j)$, the matchers estimate dense or semi-dense correspondences between local regions or pixels.
For example, MASt3R obtains correspondences through reciprocal nearest-neighbor search in descriptor space.
Geometrically verified \emph{inliers} are the subset of tentative correspondences consistent with a global two-view model estimated robustly with RANSAC.
We denote the number of such inliers by $G(r_i,s_j)$; larger values indicate stronger agreement in visible layout and local structure.
The dataset-level geometry score is likewise $\bar{G} = \tfrac{1}{N} \sum_{i=1}^{N} G_i$, where $G_i$ is the geometry score associated with the selected synthetic match for $r_i$. For SADGE and the between-/within-family decomposition, we use $\tilde{G} = \log(1 + \bar{G})$ to reduce skew; standalone correspondence results in Fig.~\ref{fig:teaser} use raw mean inlier counts $\bar{G}$.

The appearance and geometry metrics are on different scales, so we standardize them with $z$-score normalization, $\hat{A} = (\bar{A} - \mu_A)/\sigma_A$ and $\hat{G} = (\tilde{G} - \mu_G)/\sigma_G$.
Each variant yields a dataset-level pair $(\bar{A}_k,\tilde{G}_k)$ with an associated downstream score $y_k$. More precisely, each benchmark point $k \in \{1,\dots,K\}$ corresponds to a synthetic-to-real dataset comparison, i.e., one synthetic dataset variant paired with a target real dataset and its measured downstream performance under a fixed training and evaluation protocol (see Sec.~\ref{sec:results}). The statistics $(\mu_A,\sigma_A)$ and $(\mu_G,\sigma_G)$ are a \emph{single global pair}, computed once over all $K$ variants used for the exploratory fit and not per real dataset; their values are reported in the supplement. Scoring with this fitted summary reuses those stored statistics unchanged.
This normalization ensures that neither modality dominates purely due to scale, and it enables a stable joint parametrization of the final metric.

\subsection{Exploratory Joint-Signal Analysis}
\label{sec:learned_fusion}

After evaluating the existing metrics independently, we ask a secondary exploratory question: \emph{does combining appearance and geometry summarize the assembled benchmark better than either signal family alone?}
We combine normalized appearance $\hat{A}$ and geometry $\hat{G}$ using a function designed for (i) an interpretable dependence of each signal on the other, (ii) an explicit interaction between the signals, and (iii) low model complexity.
Because the inputs are $z$-scores and may be negative, monotonicity is regional rather than global: $\partial\mathrm{SADGE}/\partial\hat{G} = a + c\hat{A}$ and $\partial\mathrm{SADGE}/\partial\hat{A} = b + c\hat{G}$.
At the released coefficients the score increases in geometry once $\hat{A} > -0.623$ and in appearance once $\hat{G} > -0.588$; over the observed ranges $\hat{A} \in [-1.953, 1.150]$ and $\hat{G} \in [-1.477, 1.771]$ both conditions hold jointly for $7$ of the $21$ variants.
Agreement in one signal raises the score only where the other is not far below the benchmark mean, which is the complementarity the interaction term is designed to express.


We use a bilinear interaction so that the contribution of one signal can depend on the other.
The supplement compares it with fifteen alternative summary functions; on this benchmark the bilinear interaction is the strongest of the sixteen, ahead of the next form by $0.118$.
Specifically, we use the following interaction model:
\begin{equation}
\mathrm{SADGE} = a \hat{G} + b \hat{A} + c \hat{G}\hat{A},
\label{eq:sadge}
\end{equation}
where $a,b,c \geq 0$ are coefficients.
The linear terms represent the independent associations of geometry and appearance, while the bilinear term tests their complementarity.
This form expresses the hypothesis suggested by the metric study: appearance and structural agreement are partial descriptions of synthetic-data utility, and the usefulness of one may depend on the other.

Given a collection of benchmark datasets with known downstream performance values $\{y_k\}_{k=1}^{K}$,
we estimate $(a,b,c)$ by maximizing the Pearson correlation between predicted SADGE scores and target downstream scores:
\begin{equation}
\max_{a,b,c \geq 0} 
\ \mathrm{corr}\bigl(\{\mathrm{SADGE}_k\}_{k=1}^{K}, \{y_k\}_{k=1}^{K}\bigr).
\end{equation}
The fitted score is a descriptive synthesis of the assembled benchmark.
The pooled correlation therefore evaluates how well a compact joint summary captures the observed relationship, while the family-exclusion panels measure sensitivity to benchmark composition.


\providecommand{\bodyfigw}{1.0}
\section{Results}
\label{sec:results}

We organize the evaluation around three questions:
\textbf{Q1:} Which existing appearance- or structure-oriented metrics track synthetic-to-real downstream utility?
\textbf{Q2:} How does the answer change with the benchmark composition and the mechanism that creates the domain gap?
\textbf{Q3:} Does a compact combination of appearance and geometry summarize the pooled observations better than either family alone?
We study these questions on DIMO, VKITTI2, RarePlanes, TUD-L, ASD, Polyp-Kvasir, and Polyp-ETIS across object detection, semantic segmentation, and 6D pose estimation.
Figure~\ref{fig:datasets} illustrates all seven benchmark families; the supplementary material gives implementation details.

\subsection{Benchmark and Task Selection Protocol}
To compare metrics against a common reference, we select downstream results using a fixed protocol: (1)~we require a method trained on synthetic data and evaluated on real data (or the reverse transfer direction when defined by the source benchmark);
(2)~within each source benchmark, we select the highest reported performance under the matched protocol as an approximation of the utility available from that training dataset;
(3)~for each real test image, we calculate every appearance and geometry proxy against synthetic training images using either aligned pairs or retrieval from $k$ uniformly sampled candidates; and (4)~we aggregate each proxy per dataset variant and measure its association with the corresponding downstream result.
This protocol is designed for comparative metric analysis: within a benchmark family the training and evaluation recipe is held fixed, while the pooled analysis describes how metric behavior changes across families.

We pool task-specific downstream scores without within-task or within-family normalization.
AP, IoU, and pose average recall do not define a common utility scale, so pooled correlations are descriptive and are not invariant to family-specific rescaling.
After standardizing both proxy and downstream scores within each family, SADGE retains an in-sample pooled Pearson correlation of \(0.586\), compared with \(0.556\) for DINOv2, over 20 variants from six families.


A practical constraint is that relatively few synthetic datasets provide matched downstream evaluation on real data under controlled variants. 
Each family's downstream scores are taken from a single published table under one training and evaluation recipe, so variants within a family differ only in the synthetic source. The per-family selection rules, the source table for every value, and the transfer direction of each row are documented in the supplement and in the released variant table.

The analysis contains 21 dataset-level variants: DIMO (4 variants), VKITTI2 (6 variants), RarePlanes (1 variant), TUD-L (2 variants), ASD (2 variants), Polyp-Kvasir (3 variants), and Polyp-ETIS (3 variants). For all figures, we evaluate at most 1{,}000 test cases per variant. In retrieval mode we use $k=10$, so each query contributes 10 real--synthetic pairs. This gives the following pair counts: TUD-L (retrieval-based) 6{,}000 per variant (12{,}000 total), VKITTI2 (aligned) 1{,}000 per variant (6{,}000 total), RarePlanes (retrieval-based) 10{,}000 total, ASD (retrieval-based) 10{,}000 per variant (20{,}000 total), DIMO with 1{,}000 pairs for realpose\_reallight plus 10{,}000 pairs for each of the remaining three variants (31{,}000 total), Polyp-Kvasir (retrieval-based) 1{,}000 per variant (3{,}000 total), and Polyp-ETIS (retrieval-based) 1{,}960 per variant (5{,}880 total). In total the benchmark comprises 87{,}880 real--synthetic image-pair comparisons, but the statistical unit remains the 21 dataset variants.
The reported retrieval analysis uses one cached candidate draw shared across pairwise metrics. We assess candidate-pool size on 15 variants and candidate-pool resampling across five seeds for six polyp variants and nine proxies; these partial checks do not establish benchmark-wide seed robustness.


\subsection{What Do Existing Metrics Reveal?}
Figure~\ref{fig:teaser} reports the correlation between each metric and downstream model performance.
The first two panels use all 21 variants and report the Pearson \(r\) and the Spearman \(\rho\); the remaining panels remove one family at a time to expose sensitivity to benchmark composition.
Higher \(r\) indicates stronger linear agreement with the assembled downstream reference, and higher \(\rho\) indicates stronger agreement in ranking.
Bars are grouped by signal family: structure-oriented metrics use geometric correspondence or local structure, while appearance-oriented metrics use perceptual, pixel, distributional, or foundation-model similarity.
The complete $4\times8$ set of  joint configurations is reported in the supplement.

\textbf{Finding 1: no standalone proxy is consistently strongest across the benchmark.}
On the full benchmark the strongest appearance-only baseline is LPIPS (\(r=0.585\)), whereas the strongest structure-only baseline is LoFTR (\(r=0.545\)).
The ranking changes when the benchmark composition changes.
In particular, excluding DIMO reduces the correspondence matchers most strongly (LoFTR \(0.160\), SuperPoint+LightGlue \(0.023\), MASt3R \(-0.022\)), and the strongest standalone signals become SSIM (\(0.338\)) and LPIPS (\(0.271\)).
\textbf{Finding 2: appearance and correspondence metrics emphasize different aspects of the synthetic-to-real gap.}
Standalone metric rankings change substantially with benchmark composition.
The largest shift occurs when DIMO is excluded: the correlations of LoFTR, SuperPoint+LightGlue, and MASt3R fall to \(0.160\), \(0.023\), and \(-0.022\), respectively, while SSIM (\(0.338\)) and LPIPS (\(0.271\)) become the strongest standalone proxies.
Across the family-exclusion panels, appearance metrics tend to be more informative in settings dominated by rendering style, texture, illumination, or weather variation with approximately preserved scene layout, whereas correspondence metrics are more informative in settings involving pose, viewpoint, object placement, or spatial structure.
Because these factors are not independently controlled across families, this pattern should be interpreted as a contextual association rather than a causal attribution to individual gap mechanisms.
Nevertheless, the observed ranking instability shows that appearance and correspondence measure distinct aspects of synthetic-to-real compatibility and that neither should be treated as a universal proxy for downstream utility.

\textbf{Exploratory Finding 3: cross-signal fusion captures information not recovered by either signal family alone.}
SADGE combines DINOv3 and MASt3R, whose standalone correlations are \(r=0.232\) and \(r=0.531\), yet reaches a pooled correlation of \(r=0.834\).
The same advantage appears under the rank criterion, with \(\rho=0.853\) compared with \(0.505\) for the strongest standalone proxy.
Matched-capacity controls reach at most \(r=0.641\) when combining two appearance signals and \(r=0.558\) when combining two geometry signals, indicating that the gain is not explained by the three-parameter form alone.
Moreover, the eight strongest cross-signal configurations all use MASt3R or LoFTR, while several appearance encoders perform similarly.
The pooled result remains in-sample, but the association persists after family centering (\(r=0.524\)), and repeated variant-level held-out calibration reaches median \(r\in[0.611,0.755]\), exceeding the strongest unfitted proxy from five calibration variants onward.
Together, these results support complementary appearance and correspondence information beyond the fitted rows.


\subsection{Between- and Within-Family Decomposition}
\label{sec:results_decomposition}
A pooled correlation over the 21 variants mixes two comparisons that carry different evidential weight: differences between benchmark families, and differences between variants inside a family.
Only the second holds the training and evaluation recipe fixed, so we report the two halves separately rather than relying on the pooled number alone.
We split each score into a between-family component, the correlation across the seven family means, and a within-family component, the correlation after subtracting its family mean from every row.
Centering removes RarePlanes, which contributes no within-family variation, so the within-family estimate rests on the remaining 20 rows.

\begin{table}[t]
\centering\small
\caption{Decomposition of the pooled Pearson correlation into a between-family component (across the seven family means) and a within-family component (rows centered by family mean). Correspondence scores use $\log(1+\bar{G})$ here and in SADGE, whereas Fig.~\ref{fig:teaser} reports raw mean inlier counts: LoFTR's pooled $r$ changes from $0.545$ to $0.529$, and MASt3R's from $0.531$ to $0.457$.}
\label{tab:family_decomp}
\resizebox{\linewidth}{!}{\begin{tabular}{lccc}
\toprule
Score & pooled & between-family & within-family \\
 & ($n=21$) & (7 family means) & (20 centered rows) \\
\midrule
SADGE      & $+0.834$ & $+0.893$ & $+0.524$ \\
LPIPS      & $+0.585$ & $+0.562$ & $+0.351$ \\
log LoFTR  & $+0.529$ & $+0.677$ & $+0.235$ \\
log MASt3R & $+0.457$ & $+0.605$ & $+0.180$ \\
\bottomrule
\end{tabular}
}
\end{table}

\textbf{The joint score is the strongest on both halves of the decomposition.}
Table~\ref{tab:family_decomp} shows that the standalone matchers are carried disproportionately by between-family variation: log LoFTR falls from \(r=0.677\) between families to \(r=0.235\) within them, and log MASt3R from \(0.605\) to \(0.180\).
LPIPS is more balanced (\(0.562\) and \(0.351\)) but lower on both halves.
SADGE is the only score that stays above \(0.5\) once family means are removed (\(r=0.524\)), while also leading the between-family comparison (\(r=0.893\)).
Its pooled advantage is therefore not purely an artifact of ordering the families correctly: it retains the ranking signal that survives holding the benchmark family fixed.
This remains an in-sample, descriptive result, and the within-family estimate is based on 20 rows spanning six families, so it constrains rather than establishes generalization.

\subsection{Held-Out Calibration}
\label{sec:results_heldout}

We next examine how the amount of calibration data affects SADGE on variants excluded from parameter estimation.
The full-benchmark SADGE correlations are in-sample because the coefficients and normalization statistics are estimated on the same 21 variants that are scored.
For each calibration size \(m \in \{3,\dots,10\}\), we randomly select \(m\) variants, estimate \((\mu_G,\sigma_G,\mu_A,\sigma_A)\) and \((a,b,c)\) using only those rows, freeze all parameters, and evaluate the Pearson correlation on the remaining \(21-m\) variants.
We repeat this procedure for 2{,}000 random subsets.
We keep DINOv3 and MASt3R fixed and refit only the normalization statistics and fusion coefficients.
These splits therefore evaluate calibration of a preselected component pair, not generalization of the full component-selection procedure.
LPIPS and log LoFTR are evaluated on the same held-out rows as unfitted reference proxies.

\begin{figure}[t]
\centering
\includegraphics[width=\linewidth]{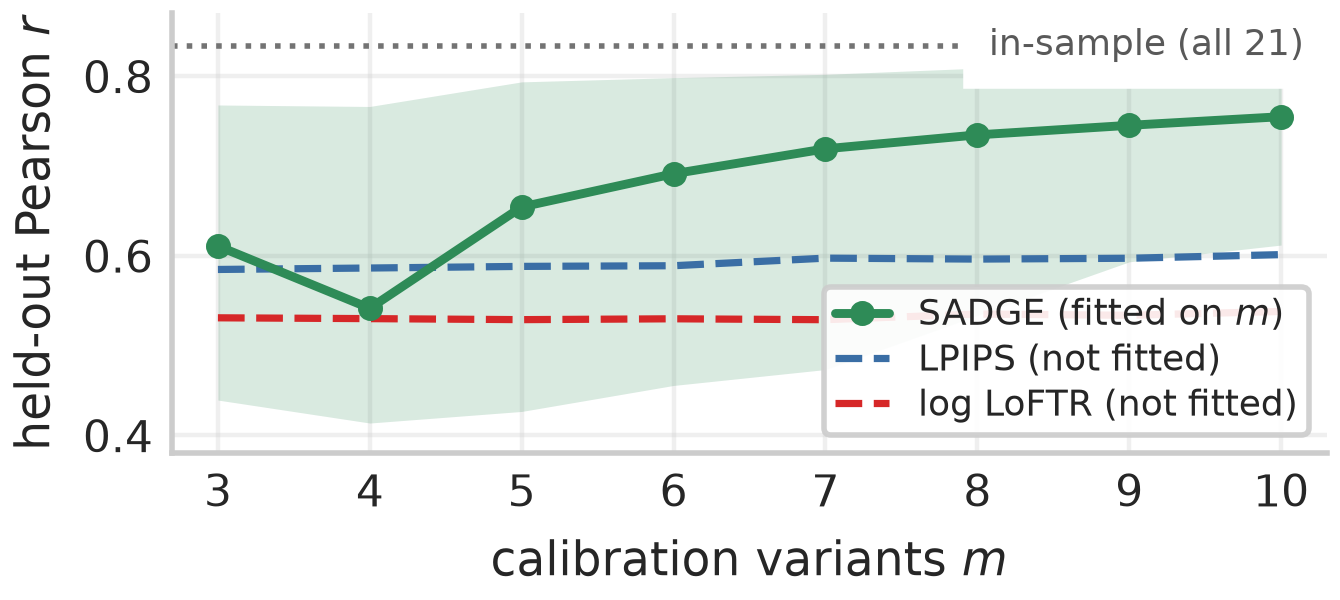}
\vspace{-6mm}
\caption{Held-out evaluation as a function of calibration-set size. Normalization and fusion coefficients are fitted on \(m\) randomly selected variants and evaluated on the remaining \(21-m\). Curves show the median and interquartile range over 2{,}000 draws; the unfitted proxies are evaluated on the same held-out rows.}
\vspace{-4mm}
\label{fig:heldout}
\end{figure}

Median held-out agreement increases from \(r=0.611\) at \(m=3\) to \(r=0.755\) at \(m=10\), whereas the unfitted proxies remain near \(0.59\) for LPIPS and \(0.53\) for log LoFTR.
SADGE exceeds the strongest unfitted proxy in median performance from \(m=5\) onward and wins on \(55\%\) of individual draws at \(m=5\), increasing to \(73\%\) at \(m=10\).
At the smallest calibration sizes the curve is non-monotonic because three coefficients can fit three calibration points exactly, yielding median calibration \(r=1.000\) but an underdetermined direction outside those points.
As \(m\) increases, this optimism decreases: the median calibration-to-held-out gap falls from \(0.39\) to \(0.11\), and calibration on ten variants recovers most of the full-sample correlation of \(0.834\).


\paragraph{Splitting whole families.}
We next ask whether calibration transfers when calibration and evaluation share no benchmark family.
For each \(f\in\{2,\dots,5\}\), we exhaustively enumerate all subsets of \(f\) calibration families, fit the normalization and fusion coefficients on their variants, and evaluate the frozen score on the remaining families.
This gives \(112\) splits, with the component pair and reference proxies fixed throughout. Median held-out correlation rises from \(r=0.529\) with two calibration families to \(0.841\) with five and exceeds both unfitted proxies once three families are available.
Although this average behavior closely follows the variant-level results, the family-disjoint splits are considerably more variable: their interquartile range is about three times wider, and \(19\%\) produce negative correlations.
Thus, the main difference is not the typical result but its dependence on which families are used for calibration.

\begin{figure}[t]
\centering
\includegraphics[width=\linewidth]{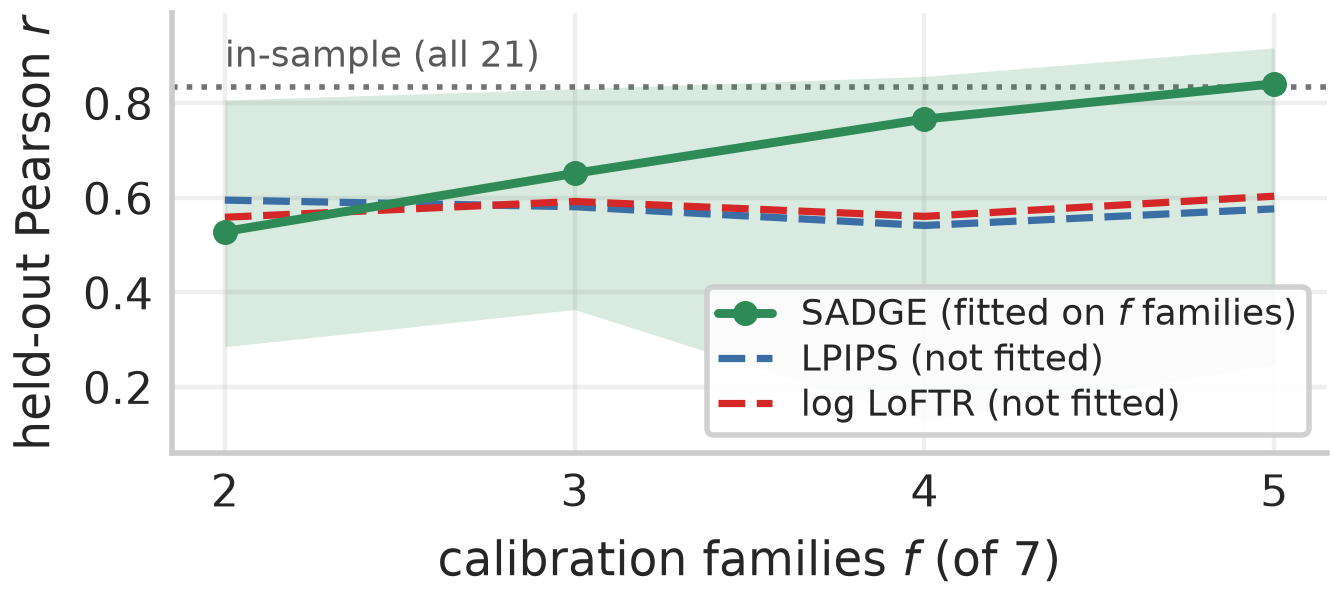}
\vspace{-6mm}
\caption{Family-disjoint calibration. The green curve shows median held-out correlation and the band the interquartile range over all exhaustive splits; unfitted proxies use the same held-out rows.}
\vspace{-4mm}
\label{fig:heldout_family}
\end{figure}

The failures follow a clear pattern.
When calibration gives negligible weight to the interaction term, median held-out correlation falls to \(-0.04\), compared with \(+0.81\) for the remaining splits.
Every negative split includes DIMO, whose variants are particularly well explained by correspondence alone and can therefore pull the fit toward a geometry-dominated solution that transfers poorly. Holding out one family at \(f=6\) reveals the same heterogeneity, with strong within-family ranking on DIMO and Polyp-ETIS but weak ranking on VKITTI2 and Polyp-Kvasir.
Overall, SADGE transfers to unseen families when calibrated on a sufficiently diverse set, but the coefficients are not universal and remain sensitive to calibration composition.

\section{Conclusion}
\label{sec:conclusion}

This work evaluates appearance- and structure-oriented metrics as proxies for synthetic-data utility.
Across seven benchmark families, 21 dataset variants, and three downstream task types, the strongest standalone proxy varies with the synthetic-to-real setting.
Appearance metrics are informative for several rendering, texture, and illumination shifts, whereas correspondence-based metrics provide stronger evidence in settings involving pose, viewpoint, or spatial structure.
The primary contribution is this empirical characterization: existing proxies have different strengths across domain-gap settings, and synthetic-data evaluation should examine both appearance and structural compatibility.

As a secondary analysis, SADGE combines these signals through a transparent three-parameter interaction.
It reaches an in-sample pooled Pearson correlation of \(r=0.834\).
Across repeated random splits, we fit SADGE on three to ten variants and evaluate it on the remaining variants.
Median held-out Pearson correlation increases from \(0.611\) to \(0.755\), exceeding the strongest unfitted proxy from five calibration variants onward.
With whole families held out, median agreement rises from \(r=0.529\) with two calibration families to \(0.841\) with five.
The interquartile range is about three times wider than for variant-level splits at a matched calibration budget; failures occur when calibration discards the interaction term.
Together, these results indicate that the benefit of combining appearance and correspondence is not confined to the full-benchmark fit.
SADGE requires calibration datasets with known downstream performance; obtaining these targets requires downstream training unless suitable published results are available.
Once calibrated, scoring additional datasets requires no downstream training, but calibration sensitivity limits reliability on unseen domains, and final downstream validation remains necessary.


\section*{Acknowledgments}
This work was supported in part by the European Union through the EOSC-ARENA project
(AI Research Enhancement through Networked Agents), Grant Agreement No.~101292597,
and by the PIAST-AI Factory project (PIAST AIF), co-financed by the EuroHPC Joint
Undertaking through the European Union's Digital Europe Programme and by Poland
as the Participating Country under Grant Agreement No.~101250730.
The authors thank Poznańskie Centrum Superkomputerowo-Sieciowe (PCSS) for providing
the computational resources used in the experiments.

\textbf{Competing interests.} The authors declare no competing interests.

{
    \small
    \phantomsection\label{refsstart}
    \bibliographystyle{ieeenat_fullname}
    \bibliography{main}
}

\end{document}